# Quantification of groundnut leaf defects using image processing algorithms

Asharf, Balasubramanian E, Sankarasrinivasan S

**Abstract**

Identification, classification, and quantification of crop defects are of paramount of interest to the farmers for preventive measures and decrease the yield loss through necessary remedial actions. Due to the vast agricultural field, manual inspection of crops is tedious and time-consuming. UAV based data collection, observation, identification, and quantification of defected leaves area are considered to be an effective solution. The present work attempts to estimate the percentage of affected groundnut leaves area across four regions of Andharapradesh using image processing techniques. The proposed method involves colour space transformation combined with thresholding technique to perform the segmentation. The calibration measures are performed during acquisition with respect to UAV capturing distance, angle and other relevant camera parameters. Finally, our method can estimate the consolidated leaves and defected area. The image analysis results across these four regions reveal that around 14 - 28% of leaves area is affected across the groundnut field and thereby yield will be diminished correspondingly. Hence, it is recommended to spray the pesticides on the affected regions alone across the field to improve the plant growth and thereby yield will be increased.

1. **Introduction:**

Image processing is considered to be the proficient and versatile tool to analyse the microscale parts of an image in many industrial and agricultural sectors [1]. The detection of plant diseases across a large scale farming area in early stage is a viable solution to predict the distress of the plants. This helps to forecast the crop production and initiate necessary remedial actions such as pesticide spraying, soil testing, seasonal crop rotation and maintaining proper water drainage. To detect the plant diseases and classify them into different stages, agriculturalists inspect them manually and make subjective decisions. However, this kind of monitoring is time-consuming, laborious, and often inaccessible to effectively cover large agricultural fields [2]. With an advancement in digital cameras and computer vision techniques, the prediction of plant disease becomes easier and efficient to achieve precision agriculture.

The agricultural monitoring systems are mainly based on the geospatial technology, automated systems and other sustainable approaches. The optimal deployment methods are often chosen based on the infrastructure and resource availability such as electricity, networking, accessibility, safety measurements and so on. As most of the agricultural fields in developing countries lack most of these facilities, more sophisticated solutions are of immediate requisite. The recent advances in aerial vehicles paved the way for finding novel solution towards agricultural monitoring especially in rural and developing areas. Specifically, the unmanned aerial vehicles (UAV) can offer a perfect solution towards agricultural monitoring for its easy transportation, independent operation and monitoring and superior manoeuvring capacity. The paper deals with deployment of UAV to monitor such remote agricultural fields and assessment scheme to obtain a maximum yield. The recent works on image based analysis on agricultural crop and irrigation planning to maximize the yield are discussed. Rastogi et. al utilized K-MEANS based segmentation, feature extraction algorithms to classify the leaf damage and develop neural network model to determine the defected leaf area [3]. Vijay et al. have classified plant leaf in different species. The classification is done with the image segmentation technique to detect the diseases automatically [4]. S.Jayalakshami et al. developed an image based diagnostic method to determine the deficiency symptoms and it is useful to the farmers for taking necessary actions earlier for increasing the yield [5]. Shanwen et al. segmented the defected leaf image with hybrid clustering. The leaf pictures are divided by superpixels clustering using Expectation Maximisation algorithm [6]. Rupali Mahajan et al. developed a genetic algorithm to classify different types of leaf disease. Speed and accuracy of detection are improved and the unhealthy leaf can be separated easily using this technique [7]. Pejman et al. introduced a scatter transform algorithm for the extraction of weed. The texture classification is performed with multiple scattering [8]. N. Hanuman et al. demonstrated to detect the diseases in microscale level and classify them into various stages of diseases at an early stage to help the farmers. The segmentation is carried out using the HIS algorithm to classify the diseases [9]. Mehmet et al. utilized the UAV to detect the defected tomatoes. The algorithm provided 90% accurate results, which helps the farmers to maximize the yield [10]. Yash et al. developed an image processing algorithm to utilize the collected RGB and thermal images for calculating the temperature of the upper part of the leaves [11]. Aditya Khamparia et al. used a hybrid convolutional neural network and autoencoder for the classification of diseases. They have used huge data set of images of leaf, detect the crop diseases by cropping them with autoencoder [12]. Poornima et. al. [13] utilized edge and color based approach to segment the diseses potion of the leaves and classified the diseases using support vector machine algorithm. Dat et.al. [14] performed segmentation, labeling, size filtering to determine the boundaries of banana leaves and extracted the color characteristics to identify the defects. Arivazhagan et.al [15] used transformation approach to convert RGB into grayscale and then applied threshold to segment the leaf defects. Singh and Misra [16] used image algorithm to segment the leaf images for automatic detection and classification of leaf diseases. Dhingra et.al. [17] discussed the various detection and classification image algorithms to identify and classify the leaf defects. Wang et.al. [18] used deep learning approach to differentiate various leaf features and defects with incorporation of VGG16, VGG19, Inception-v3, and ResNet50 models. Camargo et.al. [19] developed image algorithm to convert RGB images into H, I3a and I3b colour transformations to identify and classify the plant diseases.

In contrast to the existing approaches, the proposed method aims to perform the leaf disease detection and quantification using simple image processing method. The main advantage is that, the proposed detection method can be easily realisable in hardware and can be also deployed in futuristic UAV agricultural real time systems. In this work, groundnut crop images are collected using UAV equipped with a high definition camera across five regions of Andhra Pradesh, India. The various defects of the leaves are identified and image analysis is performed to estimate the area of defected leaf. Further, for each region group of images are selected based on their high resolution and the total leaves area is calculated. The developed image algorithm is utilized to quantify the total defected area of leaves and thus the percentage of defected leaves area is estimated.

## 2. Collection of groundnut field data using UAV

A multi-rotor UAV with a payload of 1kg and endurance of 30 minutes is utilized to collect the images using a high definition camera. The UAV is flown (Fig. 1) at an altitude of 5m with a speed of 1 m/sec in a stable manner to collect the groundnut field data. Groundnut crop field data is collected at various regions of Kadappa district such as Vedururu, Vallur, Gondipall and Ananthapur district across Rugubanpalli and Vitlampalli of Andhrapradesh

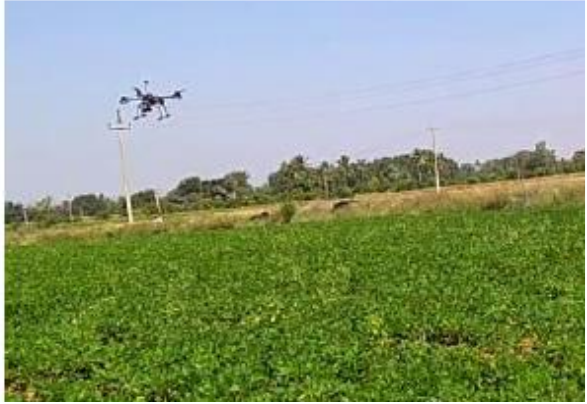

Fig.1 Collection of groundnut field data using UAV

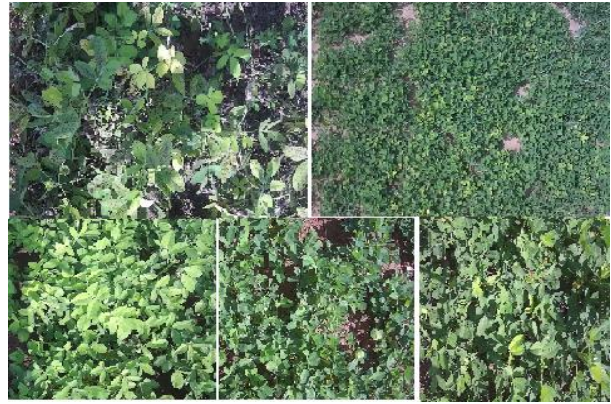

Fig. 2 Samples of collected images of groundnut leaves

## 3. Image processing and analysis of groundnut leaf

The image processing and analysis is divided into three phases as shown in Fig. 3.
- Collection of images, selection of good quality of images, and training.
- Leaf area calculation, which includes soil removing, conversion of RGB to HSV, convert gray scale into binary images
- Quantification of defected leaf areas through pixel calculations

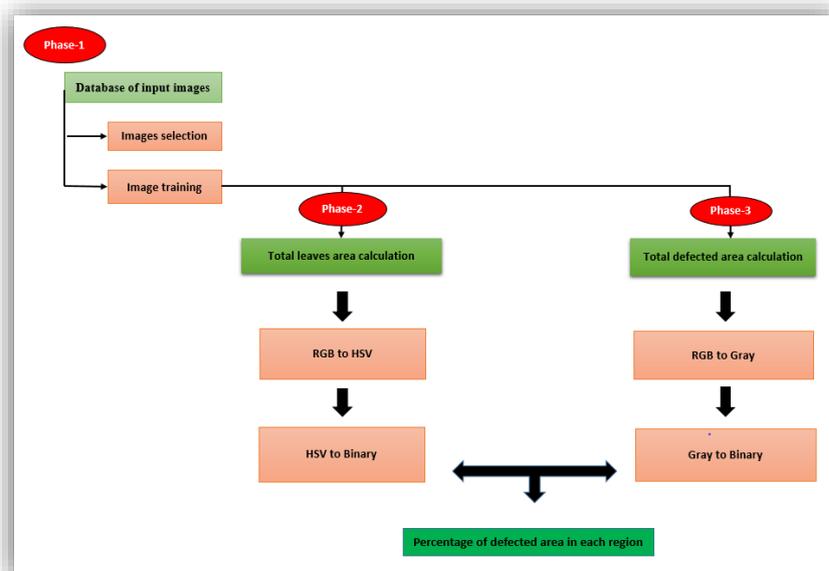

Fig. 3 Proposed approach using image processing

In order to train the defected leaves, the leaf samples with defects are considered for the analysis.

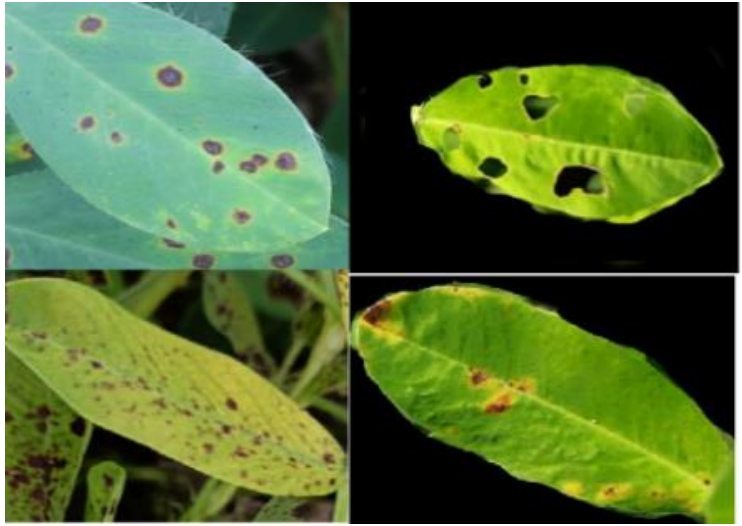

Fig.4 Leaf samples with defects

**3.1 Color Space transformation**

A. Conversion of RGB to HSV:

The collected images in the RGB scale are converted into HSV color space to remove the soil part and obtain the green part alone. The sample input and output is shown in Table 1

Table 1: Conversion of RGB to HSV

| Input | Output |
|---|---|
| 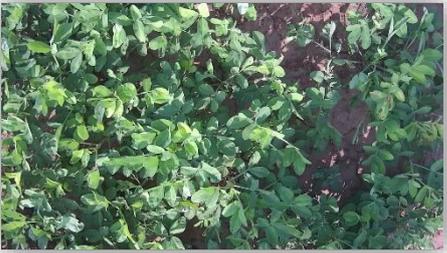 | 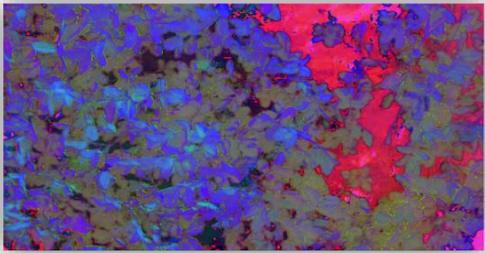 |

B. Conversion of HSV to Binary:

Further, the conversion from HSV to binary is performed to detect the green part of the leaves with a threshold value and the sample image is converted into binary as given in Table 2.

Table 2: Conversion from HSV to Binary

| Input | Output |
|---|---|
| 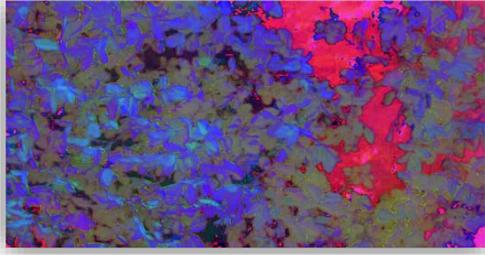 | 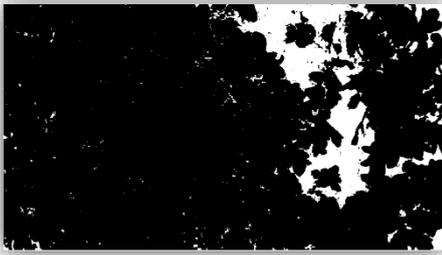 |

C. Conversion of RGB to Grayscale:

Furthermore, the transformation of images from RGB to Grayscale is performed to detect the defected part of the leaves with a threshold value as shown in Table 3. Here, for detection of defected part using the HSV scale is not possible. No unique threshold value is found to detect the defected part along with the green part.

Table 3: Conversion from RGB to Grayscale

| Input | Output |
|---|---|
| 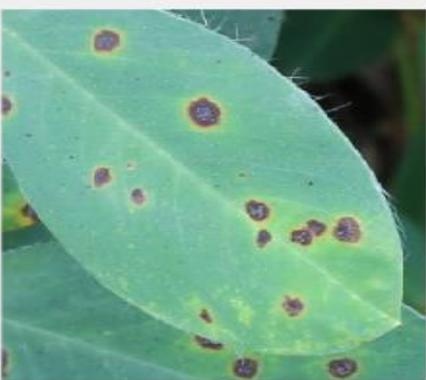 | 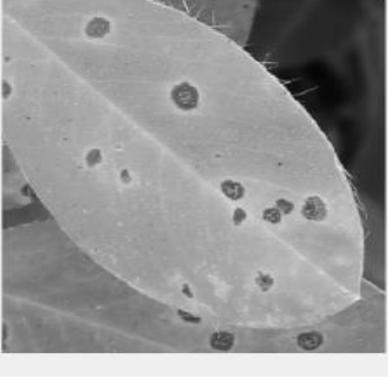 |

D. Calculation of Leaf Area

The leaf area is calculated by converting the HSV images into binary format. The total number of pixels available with respect to the green segment alone will be counted and it will be converted as a leaf area using the pixel to area conversion.

Based on the conversions from RGB to binary, the defected area of the leaf is calculated. Few samples of defected leaf are given as an input and they are converted into binary further percentage of leaf defected area is calculated and they are given in Table 4.

Table 4: Quantification of leaf defected area

| Input | Output | % of the defected area |
|---|---|---|
| 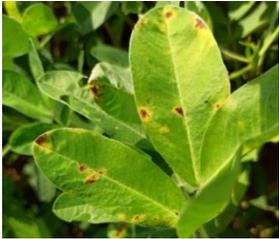 | 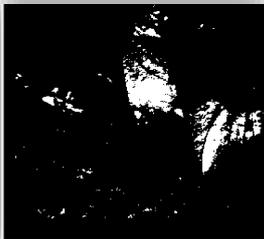 | 9.5 % |
| 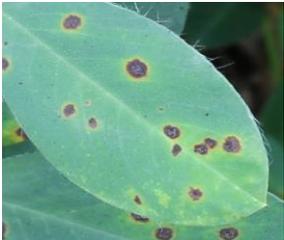 | 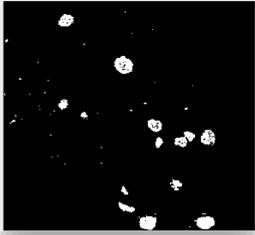 | 3.3 % |
| 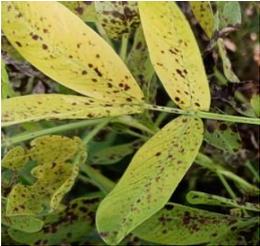 | 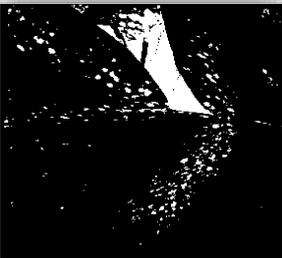 | 14.3 % |

| 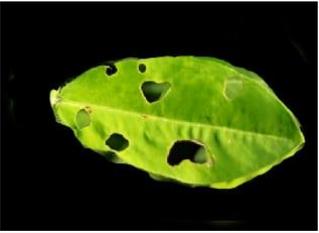 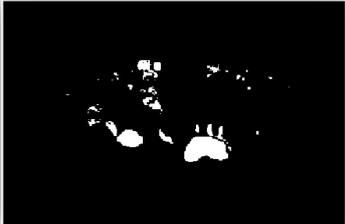 | 9.3 % |

The developed image processing algorithms are applied to the set of collected images of groundnut field data across five regions of andharapradesh to quantify the leaves defected area. In this work, high-quality images alone are considered for each region and it is varied for each region. The total leaves area in which defected leaves area is calculated and then the percentage of defect is identified. It gives an idea to the farmers that approximately 14 – 28 % of regions may get affected due to pest attacks and plant distress and correspondingly the yield will be deduced. In addition, it is observed from the Table 5 that, region Vallur has predominantly affected in comparison to other regions.

Table 5: Quantification of defected area of leaves in various regions

| Sl. No. | Region Name | Total number of images | Total Leaf Area (mm$^2$) | Total Defected Area (mm$^2$) | Percentage of Defect (%) |
|---|---|---|---|---|---|
| 1 | Vedururu | 33 | 1587.76 | 275.77 | 17.37 |
| 2 | Vallur | 32 | 1661.37 | 459.68 | 27.67 |
| 3 | Gondhipalle | 33 | 1740.76 | 258.45 | 14.85 |
| 4 | Vitlampur | 32 | 1711.21 | 389.94 | 22.79 |
| 5 | Rughubanpalli | 33 | 1842.39 | 307.45 | 19.93 |

Conclusion

UAV based collection of groundnut field data and image processing analysis is performed to quantify the defected area of groundnut leaves across five regions of Andhra Pradesh. The conversion of RGB image to HSV, grayscale, and then into binary format yielded initial segregation of removing the soil, identification of defected portions of leaves, and further calculation of total pixels into to leaf area. Image analysis considering the high-quality images alone gathered using UAV among the four regions resulted in quantifying the percentage of defected leaves area and thereby tentative estimation of yield loss is determined. It is observed that around 14 – 28 % of leaves area are affected across these regions and Vallur is affected to the maximum amongst all the regions considered for the present study. In the future, machine learning-based algorithms will be considered to estimate the defected leaves area, and precise locations across the groundnut field will be identified using UAV based georeferencing.